\documentclass{article}
\usepackage{spconf,amsmath,graphicx}
\usepackage{multicol}
\usepackage{multirow}
\usepackage{booktabs}
\usepackage{threeparttable}
\usepackage{array}
\usepackage{algpseudocode}
\usepackage{amssymb}
\usepackage{algorithm}
\usepackage{algorithmicx}
\usepackage{amsmath}
\usepackage{url}

\title{Two-Stream Multi-Task Network for Fashion Recognition}
\name{Peizhao Li $^\dag$, Yanjing Li $^\dag$, Xiaolong Jiang $^\dag$, Xiantong Zhen $^\ddag$$^\sharp$ \thanks{This work is sponsored by the National Science Foundation of China (Grant No. 61571147 and 61871016).}}
\address{$^\dag$ Beihang University, Beijing, China\\ $^\ddag$ Inception Institute of Artificial Intelligence, Abu Dhabi, UAE\\ $^\sharp$ Guangdong University of Petrochemical Technology, Guangdong, China}
\begin{document}
%
\maketitle
\begin{abstract}
In this paper, we present a two-stream multi-task network for fashion recognition. This task is challenging as fashion clothing always contain multiple attributes, which need to be predicted simultaneously for real-time industrial systems. To handle these challenges, we formulate fashion recognition into a multi-task learning problem, including landmark detection, category and attribute classifications, and solve it with the proposed deep convolutional neural network. We design two knowledge sharing strategies which enable information transfer between tasks and improve the overall performance. The proposed model achieves state-of-the-art results on large-scale fashion dataset comparing to the existing methods, which demonstrates its great effectiveness and superiority for fashion recognition.
\end{abstract}
\section{Introduction}
\label{sec:introduction}

\begin{figure*}[ht]
\centering
\includegraphics[scale=0.28]{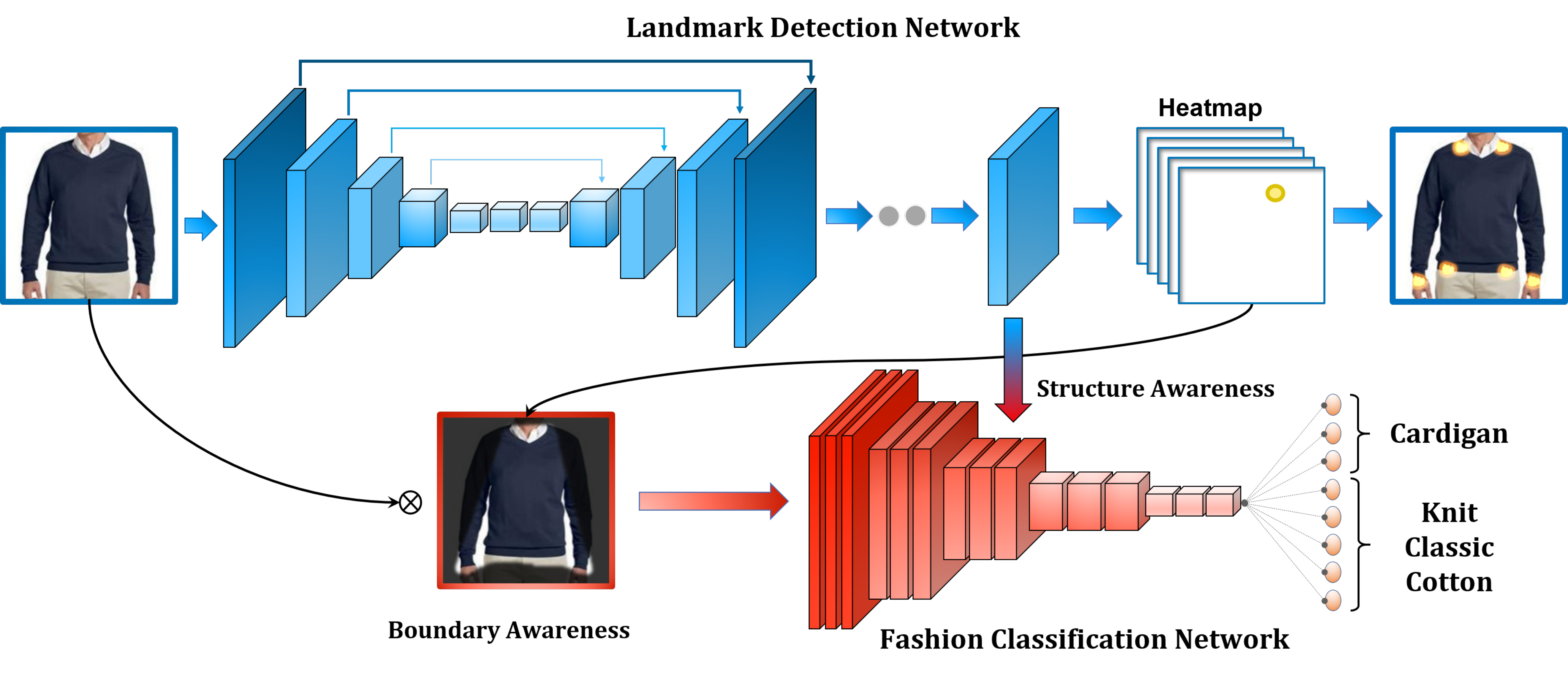}
\caption{The pipeline of the proposed framework.}
\label{fig: pipeline}
\end{figure*}

Fashion recognition has recently attracted increasing attention in computer vision for its prominent impact on electronic commerce and online shopping. However, the recognition of fashion products goes with huge challenges, with its large variations in category, style, and character as well as the deformation and occlusion, so on so forth. The images for clothes always contain rich information, such as category, attributes, and structure information. All of these need to be managed synchronously and precisely in real-time retrieval and recommendation systems, in response to the individual tastes of each user. In addition, the knowledge acquired from one kind of prediction can facilitate other predictions for multiple feature estimations. To settle the challenges of fashion prediction, we hereby formulate the problem into a multi-task learning task and solve it with the proposed end-to-end trained model.

Multi-Task Learning (MTL) \cite{mtl}, a well-known machine learning paradigm, has been successfully applied to many applications. The essence of MTL is to explore the latent connections and leverage useful knowledge between each task. By learning all tasks jointly in an MTL framework, the model can be led to achieving significant performance improvement compared with learning them individually, hence, improve the generalization performance of multiple tasks. In this paper, we manage to instantiate the power of MTL in the avenue of fashion recognition. Specifically, we integrate the tasks of landmarks detection, category classification, and attributes classification for multifarious clothes into a compact MTL framework. To solve thus formulated MTL problem, instead of a unified single-stream convolutional neural network, we design a two-stream network \cite{twostream, twostream-fusion, modelfree}, which can not only promote inter-task collaboration but also retain the capability to explore the best solution of each task individually.

However, it is non-trivial to design such a multi-task model for a specific problem. For enabling effective knowledge sharing, there are three main issues that need to be addressed: when to share, what to share, and how to share.

\textbf{When to share}. The design of CNN follows the connectivity philosophy, such that layers are stacked together in a hierarchical fashion, establishing a low-level spatial to high-level semantic information transformation. Within this hierarchy, each stage contains useful and unique information. To take full advantage of the multi-scale features reside in the rich hierarchy, we strive to enable hierarchical knowledge sharing among different tasks at both pixel-level and semantic-level across the whole network.

\textbf{What to share}. This issue is to determine the form through which knowledge sharing among all the tasks could occur. For these three tasks, we argue that the landmark detection requires a more structural feature representation, which is distinct from the category and attribute classification, leading to a heterogeneous feature distribution and parameters setting. To this end, we manage to share the structural representation for landmark detection with the classification tasks in the network, so that the structural features with rich spatial details can help better classify objects with the minute difference. Besides feature sharing, we also establish the two closely related classification tasks in a parameter-sharing way for efficacy and efficiency.

\textbf{How to share}. After the decision for when to share and what to share, we come up with two knowledge sharing methods for structural information passing, named as boundary and structure awareness. These two methods are separately implemented at pixel-level and semantic-level via boundary generation and feature aggregation, which allow the network to share knowledge in a comprehensive way.

Under the multi-task learning setting as mentioned above, we propose a two-stream convolutional neural network, wherein one branch is for landmark detection, while the other for category and attribute classification. With the purposely designed knowledge sharing strategies, this two-stream multi-task network is learned to model the correlations among multiple tasks, achieving favorable results in all tasks.

To summarise, our main contributions are threefold:

- We formulate fashion recognition into a multi-task learning problem, wherein a two-stream multi-task network is proposed to solve landmark detection, category and attribute classification collaboratively.

- We establish two concrete methods, called boundary awareness and structure awareness, for semantically representation sharing and feature aggregation among different tasks in the proposed network.

- We evaluate our approach on general benchmark dataset and achieve leading performance compared to the existing methods. Comprehensive ablation study demonstrates the contributions of each component of the framework.

\section{Related Work}
\label{sec:related work}

\textbf{Fashion Recognition} has been widely studied in recent years. As a general and important computer vision task, it composes far-reaching applications such as clothing retrieval \cite{retrieval}, recognition \cite{deepfashion}, fashion landmark detection \cite{landmark}, and clothing recommendation \cite{recommendation}. To solve this task, early methods \cite{handcrafted1} heavily rely on hand-crafted features, while recent methods mainly focus on exploiting the power of the deep neural network and have reported record-breaking results. We hereby outline some representative mile-stones for reference.

In \cite{dualattribute}, the authors leveraged a dual attribute aware mechanism for clothing retrieval. Differently, Liu \emph{et al.} \cite{deepfashion} presented a multi-branch network for clothing classification, retrieval, and landmark detection.\cite{localization} utilized a model to precisely localize attribute for fashion search. More recently, Wang \emph{et al.} \cite{attentive} proposed a compact network for landmark detection and clothing classification. Although we are similarly inspired, yet \cite{attentive} treated the landmark detection as a middle-level individual task rather than a component within a multi-task formulation. Meanwhile, this work also neglected helpful information sharing between different tasks, so that the network is more heavy with excessive parameters than ours. On the contrary, our model integrates two parameter-free approaches to accomplish information sharing among different tasks. Besides, we also manage to reduce the computational cost and enhance the adaptibility of our model.

\textbf{Multi-Task Learning} has shown promising results in many applications. A comprehensive survey can be found in \cite{survey}. In consideration of brevity, we hereby only introduce related MTL literature focusing on computer vision tasks. \cite{pose} introduced a multi-task deep convolution neural network to jointly achieve body-part and joint-point detections. In \cite{face}, a multi-linear multi-task method is proposed for person-specific facial action unit prediction. \cite{attribute} proposed a multi-task CNN for images based multi-label attribute prediction. More interestingly, \cite{rnn} presented a recurrent based framework to jointly estimate the interaction, distance, stand orientation, relative orientation, and pose estimation for immediacy prediction. Since MTL has proven its efficacy in many tasks, in this work, we are thus motivated to introduce this powerful technique into fashion recognition and implement it with the proposed two-stream multi-task network.

\section{Methodology}
\label{sec:methodology}

The overall pipeline of the proposed network is shown in Fig. 1. We denote the upper stream for landmark localization as the landmark detection network and the lower branch as the fashion classification network.

\subsection{Landmark Detection Network}
\label{sec:landmark detection network}

We build the landmark detection network following the intuition of the Hourglass structure \cite{hourglass}. As one of the most popular network for generation and reconstruction purposes, hourglass and its variants have achieved state-of-the-art results on numerous tasks including, semantic segmentation, super-resolution, and human pose estimation. For implementation, we herein deploy a stacked hourglass architecture in conjunction with intermediate supervisions, leading the network to extract robust structural representations for landmark detection. In particular, the landmark detection network is stacked with four hourglass sub-networks, inserting with bottleneck blocks for the convolution.

\subsection{Boundary Awareness}
\label{sec:boundary awareness}

The detailed operations of the boundary awareness are shown in Algorithm $1$. We construct the boundary awareness to enable knowledge sharing at the pixel-level. Given the landmark coordinates predictions, the boundary awareness automatically detect the outside edge of the target, sketch the line between the selected landmarks, and draw the boundary of the target, then generate an attention map according to the boundary segmentation in order to highlight the informative region. This generated map is finalized with a Gaussian blur. The obtained attention map can emphasize the discriminative region for clothes according to the structural information while diminishing irrelevant background. This pixel-level knowledge sharing approach works in a landmark-adaptive way, enabling to generate a fully-covered attention map, and shows robustness against abnormal landmarks distribution. Visualize illustrations for boundary awareness are provided in the ablation study.

\renewcommand{\algorithmicrequire}{$\textbf{Input:}$}  
\renewcommand{\algorithmicensure}{$\textbf{Output:}$}  

\begin{algorithm}[t]
\caption{Boundary Awareness}
	\begin{algorithmic}[1]
		\Require Image, landmarks coordinates set $\textbf P$
		\Ensure Attention map
		\State \textbf{Initialization};
	    		\State \qquad Edge set: \textbf{$\textbf E \gets \varnothing$}\;
			\State \qquad Connected landmarks set: {$\textbf C_p \gets \left\{ pt_0 \right\}$ }\;
			\State \qquad Unconnected landmarks set: {$\textbf U_p \gets \textbf P-\textbf C_p$}\;
			\State \qquad Current landmark: $pt_c \gets pt_0$\;
			\State \qquad Pseduo line between landmark i and j : $e_{ij}$\;
		\While{$\textbf U_p \neq \varnothing$}
			\For{$\forall pt_i \in\textbf U_p$ }
				\If{$\forall pt_j\in\textbf U_p+\left\{pt_0\right\}{ }$on the same side of $e_{ci}$}
					\State $E \gets E + e_{ij}$\;
					\State $pt_c \gets pt_i$\;
					\State $\textbf C_p \gets \textbf C_p + \left\{pt_i\right\}$\;
					\State $\textbf U_p \gets \textbf U_p - \left\{pt_i\right\}$\;
				\EndIf
			\EndFor
		\EndWhile
		\For{$e \in \textbf E$ }  
			\State Run Bresenham Algorithm \cite{bresenham} on $e$ $\to$ Boundary $\textbf B$\;
		\EndFor
		\State Run Scanline Fill Algorithm \cite{scanlinefill} on $\textbf B$ $\to$ attention map\;
		\State \textbf {Return} attention map
	\end{algorithmic} 
\end{algorithm}

\subsection{Structure Awareness}
\label{sec:structure awareness}

To achieve the object of knowledge sharing at the semantic level, we propose a feature map sharing method called structure awareness between the two parallel streams. In specifics, we concatenate feature maps from the end of the landmark detection network with the middle-level feature maps in fashion classification network. The feature maps from the upper stream are directly supervised and contain rich structural information. Being able to be inserted in the middle part of other tasks, these features can interact with other types of features, co-adapt and co-operate, and benefit to the downstream network for feature extraction.

\section{Experiments}
\label{sec:experiments}

\renewcommand{\arraystretch}{0.8}
\begin{table*}[tp]
  \centering
  \fontsize{9}{8}\selectfont
  \begin{threeparttable}
  \caption{Quantitative results for fashion recognition on the DeepFashion-C dataset. $\sim$ denotes result unavailable.}
  \label{tab:result}
    \begin{tabular}{p{2.5cm}<{\centering}p{3cm}<{\centering}p{2cm}<{\centering}p{2cm}<{\centering}p{2cm}	         <{\centering}p{2cm}<{\centering}}
    \toprule
    \multirow{2}{*}{Methods}&
    \multirow{2}{*}{Landmarks Detection}&
    \multicolumn{2}{c}{Category}&\multicolumn{2}{c}{Attribute}\cr
    \cmidrule(lr){3-4} \cmidrule(lr){5-6}
    &&top-3&top-5&top-3&top-5\cr
    \midrule
    WTBI \cite{handcrafted1} & $\sim$ & 43.73 & 66.26 & 27.46 & 35.37\cr
    DARN \cite{dualattribute} & $\sim$ & 59.48 & 79.58 & 42.35 & 51.95\cr
    FshionNet \cite{deepfashion} & 0.0872 & 82.58 & 90.17 & 40.52 & 54.61\cr
    DLAN \cite{dlan} & 0.0643 & $\sim$ & $\sim$ & $\sim$ & $\sim$\cr
    Corbiere \emph{el al.} \cite{corbiere} & $\sim$ & 86.30 & 92.80 & 23.10 & 30.40\cr
    Wang \emph{el al.} \cite{attentive} & 0.0484 & 90.99 & 95.78 & 51.53 & 60.95\cr
    \midrule
    Ours & \textbf{0.0467} & \textbf{93.01} & \textbf{97.01} & \textbf{59.83} & \textbf{77.91}\cr
    \bottomrule
    \end{tabular}
    \end{threeparttable}
\end{table*}

In this section, we evaluate our network on a large-scale fully-annotation fashion dataset, called DeepFashion \cite{deepfashion}. Our method substantially surpasses other methods on multiple tasks simultaneously.

\subsection{Dataset}
\label{sec:dataset}

DeepFashion is a large-scale clothes dataset with comprehensive annotations. In our experiments, we use the category and attribute prediction benchmark, which includes clothes images annotated with massive attributes, clothing landmarks, and corresponding cloth categories. The dataset consists of 209,222 images for training, 40,000 images for validation, and 40,000 images for testing, respectively.

\subsection{Implementaion}
\label{sec:implementation}

The fashion classification network is built upon an ImageNet pre-trained ResNet50. We crop the images with bounding boxes provided by the dataset and resize them into 224 $\times$ 224 for fashion classification network and 256 $\times$ 256 for landmark detection network. We replace the last two fully connected layers in fashion classification network with a two-branch fully connected layer, in which way accomplish the parameter sharing for MTL. We use mean squared error (MSE) for landmark detection, cross entropy loss for category prediction, and an asymmetric weighted cross-entropy loss for attributes estimation. We set the asymmetric weight W = 332, due to the huge positive and negative imbalance in attributes samples.

Our model is trained end-to-end with Adam optimizer on two NVIDIA TITAN X. We pre-train the landmark detection network with 10 epochs firstly, and train the two-stream network jointly with 20 epochs. Learning rate begins with 0.001, and divided by 10 for every 5 epochs.

\subsection{Quantitative Results}
\label{sec:results}

For the evaluation metric, following previous work \cite{deepfashion, attentive}, we use standard top-k classification accuracy for category classification, top-k recall rate for attribute prediction, and normalized error for landmarks detection. Our results, summarized in Table \ref{tab:result}, show that our method substantially outperforms other existing methods on multiple tasks.

\begin{figure}[ht]
\label{fig:ablation}
\centering
\includegraphics[scale=0.32]{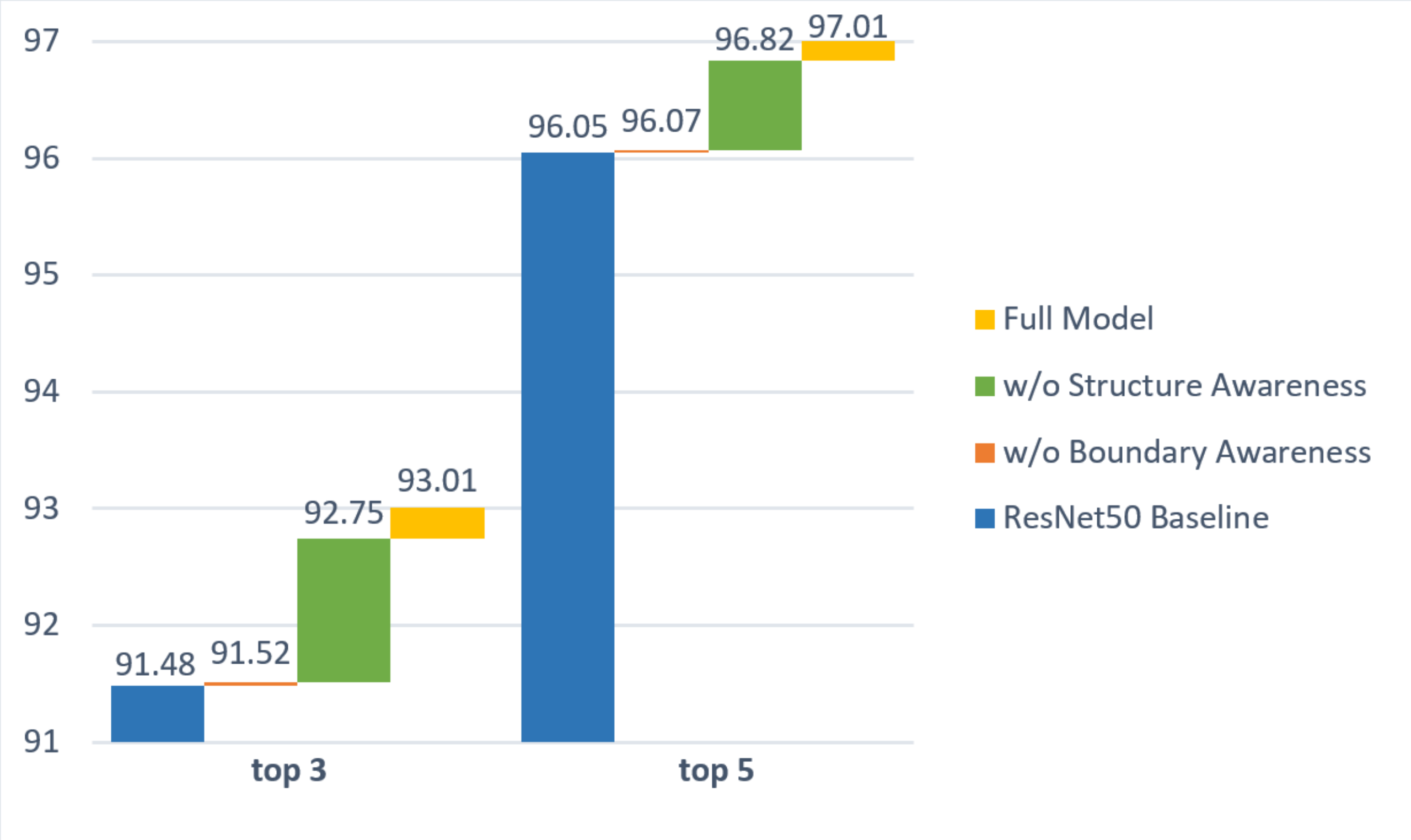}
\caption{The top-3 and top-5 category accuracy for different baseline networks.}
\end{figure}

\subsection{Ablation Study}
\label{sec:study}

Here we perform a full ablation study to evaluate the importance of all the proposed components in our model. We disable each awareness method and build baseline networks for comparison. Fig. 2 shows the category prediction score on the DeepFashion dataset of each baseline approach in comparison with the full model. The quantitative results show that each of the module contributes to the overall performance, thereby demonstrating the effectiveness of multi-task learning and the proposed awareness methods.

We do have conducted some experiments on the bidirectional message passing between the parallel network, and we observe that the feature sharing to the landmark detection network hurt the performance of landmark localization. We argue that landmark detection only requires the structural feature that represents the overall structure of the clothes, regardless of the detail feature. Sharing the feature from the classification stream may confuse the landmark detection network, leading to the performance distortion. This architecture can be formulated more as the asymmetrical multi-task learning.

We also provide some visualization result for the attention map generated by our boundary awareness in Fig. 3, compared with the fashion attention mechanism used in Wang \emph{et al.} \cite{attentive}. Being able to benefit directly from the landmark detection, the boundary awareness transports structural representation and provides a fully-covered attention map for the classification task.

\begin{figure}[ht]
\label{fig:visualization}
\centering
\includegraphics[scale=0.55]{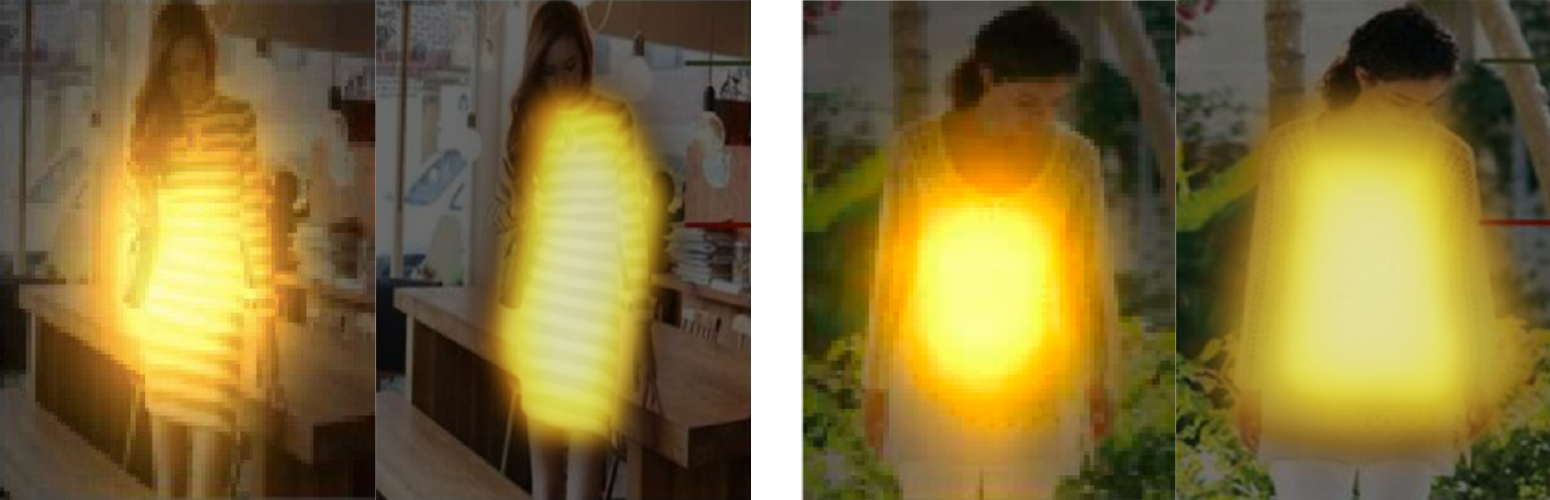}
\caption{The visualization result for attention map. For each pair of the visualization, the left one is from Wang \emph{et al.} \cite{attentive}, while the right one is ours.}
\end{figure}

\section{Conclusions}
\label{sec:conclusions}

In this paper, we present a two-stream multi-task network for fashion recognition. Our model integrated the proposed knowledge sharing methods and outperforms previous methods on fashion landmark detection, category prediction, and attribute estimation. The method can also be extended to other challenging multi-task problems and is also applicable to many computer vision related tasks.

\bibliographystyle{IEEEbib}
\bibliography{strings,refs}

\begin{thebibliography}{10}

\bibitem{mtl}
Rich Caruana,
\newblock ``Multitask learning,''
\newblock {\em Machine learning}, vol. 28, no. 1, pp. 41--75, 1997.

\bibitem{twostream}
Karen Simonyan and Andrew Zisserman,
\newblock ``Two-stream convolutional networks for action recognition in
  videos,''
\newblock in {\em Advances in neural information processing systems}, 2014, pp.
  568--576.

\bibitem{twostream-fusion}
Christoph Feichtenhofer, Axel Pinz, and Andrew Zisserman,
\newblock ``Convolutional two-stream network fusion for video action
  recognition,''
\newblock in {\em Proceedings of the IEEE Conference on Computer Vision and
  Pattern Recognition}, 2016, pp. 1933--1941.

\bibitem{modelfree}
Xiaolong Jiang, Peizhao Li, Xiantong Zhen, and Xianbin Cao,
\newblock ``Model-free tracking with deep appearance and motion features
  integration,''
\newblock {\em arXiv preprint arXiv:1812.06418}, 2018.

\bibitem{retrieval}
M~Hadi~Kiapour, Xufeng Han, Svetlana Lazebnik, Alexander~C Berg, and Tamara~L
  Berg,
\newblock ``Where to buy it: Matching street clothing photos in online shops,''
\newblock in {\em Proceedings of the IEEE international conference on computer
  vision}, 2015, pp. 3343--3351.

\bibitem{deepfashion}
Ziwei Liu, Ping Luo, Shi Qiu, Xiaogang Wang, and Xiaoou Tang,
\newblock ``Deepfashion: Powering robust clothes recognition and retrieval with
  rich annotations,''
\newblock in {\em Proceedings of the IEEE conference on computer vision and
  pattern recognition}, 2016, pp. 1096--1104.

\bibitem{landmark}
Ziwei Liu, Sijie Yan, Ping Luo, Xiaogang Wang, and Xiaoou Tang,
\newblock ``Fashion landmark detection in the wild,''
\newblock in {\em European Conference on Computer Vision}. Springer, 2016, pp.
  229--245.

\bibitem{recommendation}
Xintong Han, Zuxuan Wu, Yu-Gang Jiang, and Larry~S Davis,
\newblock ``Learning fashion compatibility with bidirectional lstms,''
\newblock in {\em Proceedings of the 2017 ACM on Multimedia Conference}. ACM,
  2017, pp. 1078--1086.

\bibitem{handcrafted1}
Huizhong Chen, Andrew Gallagher, and Bernd Girod,
\newblock ``Describing clothing by semantic attributes,''
\newblock in {\em European conference on computer vision}. Springer, 2012, pp.
  609--623.

\bibitem{dualattribute}
Junshi Huang, Rogerio~S Feris, Qiang Chen, and Shuicheng Yan,
\newblock ``Cross-domain image retrieval with a dual attribute-aware ranking
  network,''
\newblock in {\em Proceedings of the IEEE international conference on computer
  vision}, 2015, pp. 1062--1070.

\bibitem{localization}
Kenan~E Ak, Ashraf~A Kassim, Joo~Hwee Lim, and Jo~Yew Tham,
\newblock ``Learning attribute representations with localization for flexible
  fashion search,''
\newblock in {\em Proceedings of the IEEE Conference on Computer Vision and
  Pattern Recognition}, 2018, pp. 7708--7717.

\bibitem{attentive}
Wenguan Wang, Yuanlu Xu, Jianbing Shen, and Song-Chun Zhu,
\newblock ``Attentive fashion grammar network for fashion landmark detection
  and clothing category classification,''
\newblock in {\em Proceedings of the IEEE Conference on Computer Vision and
  Pattern Recognition}, 2018, pp. 4271--4280.

\bibitem{survey}
Yu~Zhang and Qiang Yang,
\newblock ``A survey on multi-task learning,''
\newblock {\em arXiv preprint arXiv:1707.08114}, 2017.

\bibitem{pose}
Sijin Li, Zhi-Qiang Liu, and Antoni~B Chan,
\newblock ``Heterogeneous multi-task learning for human pose estimation with
  deep convolutional neural network,''
\newblock in {\em Proceedings of the IEEE conference on computer vision and
  pattern recognition workshops}, 2014, pp. 482--489.

\bibitem{face}
Timur Almaev, Brais Martinez, and Michel Valstar,
\newblock ``Learning to transfer: transferring latent task structures and its
  application to person-specific facial action unit detection,''
\newblock in {\em Proceedings of the IEEE International Conference on Computer
  Vision}, 2015, pp. 3774--3782.

\bibitem{attribute}
Abrar~H Abdulnabi, Gang Wang, Jiwen Lu, and Kui Jia,
\newblock ``Multi-task cnn model for attribute prediction,''
\newblock {\em IEEE Transactions on Multimedia}, vol. 17, no. 11, pp.
  1949--1959, 2015.

\bibitem{rnn}
Xiao Chu, Wanli Ouyang, Wei Yang, and Xiaogang Wang,
\newblock ``Multi-task recurrent neural network for immediacy prediction,''
\newblock in {\em Proceedings of the IEEE international conference on computer
  vision}, 2015, pp. 3352--3360.

\bibitem{hourglass}
Alejandro Newell, Kaiyu Yang, and Jia Deng,
\newblock ``Stacked hourglass networks for human pose estimation,''
\newblock in {\em European Conference on Computer Vision}. Springer, 2016, pp.
  483--499.

\bibitem{bresenham}
Michael L.~V. Pitteway and Dereck~J Watkinson,
\newblock ``Bresenham's algorithm with grey scale,''
\newblock {\em Communications of the ACM}, vol. 23, no. 11, pp. 625--626, 1980.

\bibitem{scanlinefill}
``Scanline fill algorithm,''
  \url{http://web.cs.ucdavis.edu/~ma/ECS175_S00/Notes/0411_b.pdf}.

\bibitem{dlan}
Sijie Yan, Ziwei Liu, Ping Luo, Shi Qiu, Xiaogang Wang, and Xiaoou Tang,
\newblock ``Unconstrained fashion landmark detection via hierarchical recurrent
  transformer networks,''
\newblock in {\em Proceedings of the 2017 ACM on Multimedia Conference}. ACM,
  2017, pp. 172--180.

\bibitem{corbiere}
Charles Corbiere, Hedi Ben-Younes, Alexandre Ram{\'e}, and Charles Ollion,
\newblock ``Leveraging weakly annotated data for fashion image retrieval and
  label prediction,''
\newblock in {\em Computer Vision Workshop (ICCVW), 2017 IEEE International
  Conference on}. IEEE, 2017, pp. 2268--2274.

\end{thebibliography}

\end{document}